\documentclass{article}

\usepackage{microtype}
\usepackage{graphicx}
\usepackage{subfigure}
\usepackage{booktabs} 
\usepackage{amsmath,amsfonts,amssymb}
\usepackage{algorithm2e}
\usepackage{mathtools}
\usepackage{bm} 
\usepackage[normalem]{ulem}
\useunder{\uline}{\ul}{}
\usepackage{array}      

\usepackage{hyperref}

\usepackage{array}
\usepackage[svgnames, table]{xcolor}
\usepackage{tabularx, makecell, linegoal}
\definecolor{LightGreen}{rgb}{0.9,1.0,0.9} 
\usepackage{blindtext}
\usepackage[most]{tcolorbox}
\makeatletter
\NewDocumentCommand{\mynote}{+O{}+m}{%
  \begingroup
  \tcbset{%
    noteshift/.store in=\mynote@shift,
    noteshift=1.5cm
  }
  \begin{tcolorbox}[nobeforeafter,
    enhanced,
    sharp corners,
    toprule=1pt,
    bottomrule=1pt,
    leftrule=0pt,
    rightrule=0pt,
    colback=WhiteSmoke!80!Lavender,
    #1,
    left skip=\mynote@shift,
    right skip=\mynote@shift,
    overlay={\node[right] (mynotenode) at ([xshift=-\mynote@shift]frame.west) {\textbf{Note:}} ;},
    ]
    #2
  \end{tcolorbox}
  \endgroup
  }
\makeatother

\usepackage[accepted]{icml2025}

\usepackage{amsmath}
\usepackage{amssymb}
\usepackage{mathtools}
\usepackage{amsthm}
\usepackage{graphicx}
\usepackage{booktabs}
\usepackage{multirow}

\usepackage[capitalize,noabbrev]{cleveref}

\theoremstyle{plain}

\theoremstyle{definition}

\theoremstyle{remark}

\DeclarePairedDelimiterX{\norm}[1]{\lVert}{\rVert}{#1}

\usepackage[textsize=tiny]{todonotes}

\begin{document}

\onecolumn  

\makeatletter
\renewcommand{\twocolumn}[1][]{#1}  
\makeatother




\icmltitle{Cross-Modal Diffusion for Biomechanical Dynamical Systems Through Local Manifold Alignment}


\begin{icmlauthorlist}
\icmlauthor{Sharmita Dey}{}\footnotemark[1]
\icmlauthor{Sarath Ravindran Nair}{}\footnotemark[2]
\end{icmlauthorlist}


\footnotetext[1]{\textbf{Sharmita Dey} \\ 
\textit{ETH Zurich, Switzerland; \\ University of Göttingen, Germany} \\ 
Corresponding author: \texttt{contact.deysharmita@gmail.com} \\ 
Sharmita Dey conceptualized and led the project, developed the model, conducted the experiments, performed the analyses, and wrote and revised the paper.}  

\footnotetext[2]{\textbf{Sarath Ravindran Nair} \\ 
\textit{ENI Göttingen, Germany} \\ 
Sarath Ravindran Nair contributed to performing the analyses, participated in discussions, and co-authored and revised the paper.}


\icmlkeywords{Machine Learning, ICML}





\begin{abstract}

We present a mutually aligned diffusion framework for cross-modal biomechanical motion generation, guided by a dynamical systems perspective. By treating each modality, e.g., observed joint angles ($X$) and ground reaction forces ($Y$), as complementary observations of a shared underlying locomotor dynamical system, our method aligns latent representations at each diffusion step, so that one modality can help denoise and disambiguate the other. Our alignment approach is motivated by the fact that local time windows of $X$ and $Y$ represent the same phase of an underlying dynamical system, thereby benefiting from a shared latent manifold. We introduce a simple local latent manifold alignment (LLMA) strategy that incorporates first-order and second-order alignment within the latent space for robust cross-modal biomechanical generation without bells and whistles. Through experiments on multimodal human biomechanics data, we show that aligning local latent dynamics across modalities improves generation fidelity and yields better representations.


\end{abstract}

\section{Introduction}
\label{submission}

Many physical and biological processes exhibit multiple data streams that stem from a shared underlying process, yet capture different facets of the global dynamics \cite{ren2022predictable,ashe2006interpretation}. Biomechanical motion is a prime example of such a multimodal dynamical system: various physical quantities, such as joint angles, joint moments, and ground reaction forces (GRFs) offer complementary perspectives on the same underlying locomotor system. The inherent synergy in human locomotion further posits that the different modalities that describe the motion are correlated to each other \cite{winter2009biomechanics}. Understanding how these complementary modalities co-evolve and influence each other is useful for gaining a holistic view of human movement and other complex motor patterns.

Despite the potential richness of complementary modalities, real-world constraints often limit which modalities can be reliably measured simultaneously. High equipment costs, specialized laboratory environments, and user discomfort can make fully instrumented setups impractical, particularly for long-term or at-home monitoring. Moreover, sensor malfunctions, intermittent connectivity, data dropouts, and sensor failures often leave us with partial and noisy observations of the underlying dynamical process. Under such constraints, a cross-modal biomechanical generation framework can leverage the available observations to infer the missing modalities, such as estimating GRFs from joint kinematics alone, enabling richer analysis without relying on complex sensory setup.

Motivated by these considerations, we aim to leverage the complementary information between modalities for designing a cross-modal diffusion framework that can facilitate bidirectional biomechanical generation and inference. To achieve this, we adopt a dynamical systems perspective, positing that multiple observational streams originate from a single, time-evolving hidden state. Specifically, modalities such as joint angles ($X$) and GRFs ($Y$) can be considered as distinct observations of an underlying dynamical process governing locomotion. This perspective motivates the alignment of their latent representations within a shared manifold, thus leveraging information from each partial observation to enhance the robustness and fidelity of cross-modal generation.

\begin{figure*}
    \centering
    \includegraphics[width=\linewidth]{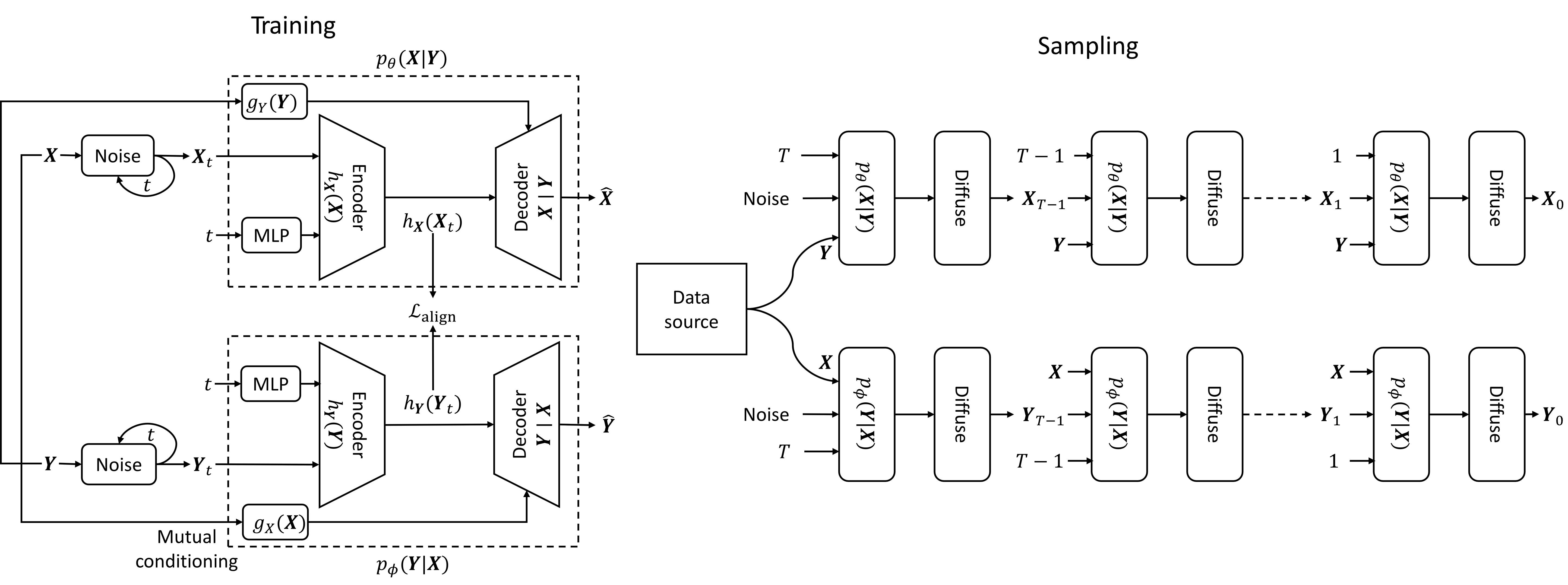} 
    \caption{(Left) Mutually aligned cross-modal diffusion with latent manifold alignment. Diffusion processes, $p_\theta(\mathbf{X}|\mathbf{Y})$ and $p_\phi(\mathbf{Y}|\mathbf{X})$, generate data for modalities, $\mathbf{X}$ and $\mathbf{Y}$, respectively, guided by a condition derived from the other modality. During training, the latent representations, $h_X(\mathbf{X}_t, t)$ and $h_Y(\mathbf{Y}_t, t)$, of the two models are aligned using a local latent manifold alignment (LLMA) objective. Additionally, denoising and energy conservation objectives are applied to each modality's generated samples, $\hat{\mathbf{X}}$ and $\hat{\mathbf{Y}}$. During sampling, the model for each modality diffuses a noise signal across $T$ timesteps, guided by a condition from the other modality to generate samples of a given modality that temporally corresponds to the guiding signal. }
    \label{fig:architecture}
\end{figure*}

We introduce a mutually aligned cross-modal diffusion framework (Fig. \ref{fig:architecture}) that couples the conditioned diffusion processes, $p({X}|{Y})$ and $p({Y}|{X})$ of modalities, $X$ and $Y$, by aligning their latents at each diffusion time step. 
To align the modalities, we introduce a local latent manifold alignment (LLMA) strategy that incorporates first-order sequence-contrastive and second-order covariance alignment within the latent space. This approach enforces alignment such that the local neighborhoods of each modality in the latent space are not only matched in their immediate representations but also preserve consistent internal correlation structures. 

To the best of our knowledge, this is the first study demonstrating cross-modal biomechanical diffusion with latent alignment grounded in a dynamical systems perspective. To summarize, our contributions are as follows: 
\begin{enumerate}
    \item We introduce a mutually aligned diffusion framework for cross-modal biomechanical time-series generation through latent representation alignment.
    \item We propose a local latent manifold alignment strategy grounded in dynamical systems principles for aligning the latent representations of the modalities.
    \item We demonstrate through experiments that this simple latent alignment strategy not only enhances generative quality but also maintains robust representations for downstream discriminative tasks.
\end{enumerate}

\section{Related Work}

\subsection{Diffusion models}

Denoising diffusion probabilistic models (DDPMs) have emerged as powerful generative frameworks for high-dimensional data. \cite{pmlr-v37-sohl-dickstein15} introduced the foundational concept of diffusion models, which was further refined by  \cite{ho2020denoising} to improve scalability and performance in image generation tasks \cite{dhariwal2021diffusion}. Extensions of diffusion models to time-series data \cite{yuan2024diffusion, shen2023non} have shown promise in areas such as speech synthesis \cite{kong2020diffwave, chen2020wavegrad} and time series forecasting \cite{kollovieh2024predict}, and anomaly prediction \cite{xiao2023imputation}.  
However, the application of diffusion models for cross-modal generation of time-series observations remains limited. 

\subsection{Cross-modal learning}

Cross-modal learning enables data in one modality to be synthesized from or interpreted by another, leveraging the complementary nature of diverse data types such as text, images, audio, and video. Landmark successes in text-to-image generation, exemplified by DALL-E \cite{ramesh2021zero,ramesh2022hierarchical} and CLIP \cite{radford2021learning}, illustrate how large-scale multimodal training can produce coherent visual outputs conditioned on textual prompts. Beyond text-to-image, cross-modal techniques now encompass music-to-dance \cite{tseng2023edge,zhuang2022music2dance}, text-to-video \cite{blattmann2023align}, text-to-motion \cite{tevet2023human}, and audio-visual scene understanding \cite{alamri2019audio}.
Despite these advances, most studies have focused on \emph{unidirectional} mappings, e.g., from text to images, without learning the corresponding inverse direction. As a result, they cannot fully capture the bidirectional nature of many real-world relationships, which can be crucial in applications that require switching between modalities. Moreover, applying cross-modal strategies to physiological time-series data, such as biomechanics, has received limited attention. Such domains require methods that can handle continuous, temporal signals and preserve the underlying dynamical structure across modalities, highlighting a need for bidirectional cross-modal models designed specifically for time-series data.

\subsection{Representation alignment}
Representation alignment is the process of mapping diverse inputs into a shared latent space that preserves structural and semantic information. 
Recent advances in learning-based representation alignment have significantly shaped the field of self-supervised learning, with prominent methods such as SimCLR \cite{chen2020simple}, Barlow Twins \cite{zbontar2021barlow}, and VICReg \cite{bardes2021vicreg} yielding impressive gains on a range of discriminative downstream tasks. These approaches excel at learning invariant and robust representations in unimodal scenarios. However, these methods have primarily been used as pretraining strategies, rather than an active process within diffusion-based frameworks. Moreover, these methods are not explicitly designed to consider the temporal structure and correlations of different modalities that are essential for coherent cross-modal generation in time series applications. By contrast, our work leverages latent representation alignment based on a dynamical systems perspective as a framework for cross-modal biomechanical time series generation.

\subsection{Biomechanical motion analysis and synthesis}

Human motion analysis integrates kinematic data, like joint angles, with dynamic data, such as ground reaction forces (GRFs). Foundational studies, such as Winter’s work on biomechanics, highlight the interplay between kinematics and dynamics in locomotion \cite{winter2009biomechanics}. Recent learning-based models have enhanced tasks like motion estimation \cite{halilaj2018machine, gurchiek2019estimating, horst2023modeling, dey2019support, dey2020continuous} but often rely on handcrafted features and single-modal datasets, limiting robustness and generalization. Fusion methods, combining data like motion capture and electromyography (EMG), improve predictions of muscle forces and joint dynamics \cite{sartori2012estimation, young2014analysis}. However, cross-modal synthesis of biomechanical motion patterns remains severely underexplored, restricting research to predefined scenarios and subject-specific solutions.

Our work addresses these challenges by introducing a cross-modal generation method specifically designed for biomechanical time series data, leveraging latent representation alignment and principles from dynamical systems. While the proposed approach is broadly applicable to various time series domains, we focus on biomechanical data as an illustrative example. This focus is intentional, as biomechanical modalities such as joint angles, joint moments, and ground reaction forces share an underlying dynamical process, a key characteristic that our method effectively utilizes to achieve robust and accurate cross-modal generation. 

\section{Cross-modal Denoising Diffusion with  Latent Alignment}

\subsection{Problem formulation}
\label{sec:problem_formulation}

We consider a dataset $\{(\mathbf{X}_i, \mathbf{Y}_i)\}_{i=1}^N$ of paired time series from distinct but related modalities (e.g., joint angles vs. joint moments, or angles vs. ground reaction forces). Each $\mathbf{X}_i \in \mathbb{R}^{L \times d_X}$ and $\mathbf{Y}_i \in \mathbb{R}^{L \times d_Y}$ is a sequence of length $L$, with $d_X$ and $d_Y$ denoting the dimensionalities of the respective modalities. Our goal is to learn generative models $p_\theta(\mathbf{X}\mid \mathbf{Y})$ and $p_\phi(\mathbf{Y}\mid \mathbf{X})$ ($\theta$ and $\phi$ are model parameters) such that one modality can be generated or reconstructed conditioned on the other.

\subsection{Denoising diffusion}

We adopt a denoising diffusion framework to learn these cross-modal distributions. Let $\beta_t$ for $t=1,\dots,T$ define a noise schedule that controls the noise variance at each step $t$ of the diffusion process. We define the following forward noising processes for each modality:
\begin{equation}
    \begin{aligned}
    \mathbf{X}_t &=  \sqrt{\beta_t}\,\mathbf{X}_{t-1} \;+\; \sqrt{1-\beta_t}\,\boldsymbol{\epsilon},\\
    \mathbf{Y}_t &=  \sqrt{\beta_t}\,\mathbf{Y}_{t-1} \;+\; \sqrt{1-\beta_t}\,\boldsymbol{\epsilon},
    \end{aligned}
\end{equation}
where 
$\boldsymbol{\epsilon}\sim \mathcal{N}(0,\,\mathbf{I})$ is standard Gaussian noise. 

We model the reverse process using conditional denoising diffusion processes, which predict the clean signal based on the noisy sample at each time step, $t$, and a condition derived from the other modality:
\begin{equation}
    \begin{aligned}
        p_\theta(\mathbf{X}_0 \;\mid\; \mathbf{X}_t,\; g_Y(\mathbf{Y}),\; t), \\
        p_\phi(\mathbf{Y}_0 \;\mid\; \mathbf{Y}_t,\; g_X(\mathbf{X}),\; t),
    \end{aligned}
\end{equation}
where $\theta$ and $\phi$ are parameters of the diffusion models, and $g_X(\cdot)$, $g_Y(\cdot)$ denote condition embedding functions for the modalities $\mathbf{X}$ and $\mathbf{Y}$, respectively.

We incorporate a mutual conditioning mechanism such that the generation of one modality is guided by the latent or encoded features from the other modality. Concretely, this means each decoder attends to both the noisy embedding of its own modality at time $t$ and a learned condition embedding derived from the other modality. 
For learning robust cross-modal representations, we enforce an alignment of the latent representations of the two modalities at each diffusion step. Since our modalities represent time-series data, we propose a modified alignment to ensure the temporal correlation of the local dynamics of the two modalities.

\subsection{Latent alignment with diffusion}

\paragraph{Dynamical systems background.} In biomechanics, two modalities, e.g., joint angles $\mathbf{X}\in \mathbb{R}^{L\times d_X}$ and joint moments $\mathbf{Y}\in \mathbb{R}^{L\times d_Y}$ can be seen as \emph{observational streams} of the \emph{same underlying dynamical system} since they stem from the same musculoskeletal control process. Formally, consider a (possibly high-dimensional) hidden state, \(\textbf{Z}\in \mathbb{R}^{L\times d_Z}\) evolving according to an unknown dynamics: 

\[
    \mathbf{Z}_{k+1} \;=\; f(\mathbf{Z}_k) \;+\; \boldsymbol{\eta}_k,
\]
where $\boldsymbol{\eta}_k$ is a noise term. The observation functions, $ o_\mathbf{X}$ and $o_\mathbf{Y}$, map the latent state into each modality's domain:
\[
    \mathbf{X}_k = o_\mathbf{X}(\mathbf{Z}_k), 
    \quad
    \mathbf{Y}_k = o_\mathbf{Y}(\mathbf{Z}_k).
\]

Under this perspective, $\mathbf{X}_k$ and $\mathbf{Y}_k$ arise from the same $\mathbf{Z}_k$ and thus should lie on correlated sub-manifolds of the global dynamical system. From Takens' embedding theorem~\cite{takens2006detecting} and related results in nonlinear time-series analysis \cite{sauer1991embedology}, such partial views can still reconstruct consistent attractors or trajectories in phase space if appropriately embedded. This perspective underlies the motivation for aligning $\mathbf{X}$-space and $\mathbf{Y}$-space: if they come from the same dynamical manifold, then local segments of the latent dynamics should describe \textit{the same underlying phase} and \textit{the same local trajectories} (up to a smooth invertible transform).

\paragraph{Local latent manifold alignment (LLMA).} In our \emph{mutually-aligned diffusion} approach, we train the diffusion models $p(\mathbf{X}\mid \mathbf{Y})$ and $p(\mathbf{Y}\mid \mathbf{X})$ simultaneously to reconstruct the modalities, $\textbf{X}$ and $\textbf{Y}$, conditioned on each other. At each timestep $k$, the diffusion models produce latent embeddings, $\textbf{Z}_\textbf{X}(k) \in \mathbb{R}^{L\times d_{Z}}$ and $\textbf{Z}_\textbf{Y}(k) \in \mathbb{R}^{L\times d_{Z}}$. From a dynamical systems perspective, we may consider these latent embeddings as a reconstruction of the local phase space of the underlying dynamical system from each sensor`s noisy observations. Since $\textbf{Z}_\textbf{X}(k)$ and $\textbf{Z}_\textbf{Y}(k)$ are reconstructions of the same underlying trajectory $\textbf{Z}_k$, they should be aligned to each other. 

We partition the latent sequences from the two models, $\textbf{Z}_\textbf{X}$ and $\textbf{Z}_\textbf{Y}$ into  subsequences of length $C$, 

\begin{equation}
    \begin{aligned}
        \textbf{Z}_\textbf{X} &= h_{\mathbf{X}}(\mathbf{X}) = [\textbf{Z}_\textbf{X}^{(1)}, \textbf{Z}_\textbf{X}^{(2)},.., \textbf{Z}_\textbf{X}^{(M)}],\\
        \textbf{Z}_\textbf{Y} &= h_{\mathbf{Y}}(\mathbf{Y}) = [\textbf{Z}_\textbf{Y}^{(1)}, \textbf{Z}_\textbf{Y}^{(2)},.., \textbf{Z}_\textbf{Y}^{(M)}], 
    \end{aligned}
\end{equation}

where $M\approx L/C$. For each index $i=1,\dots,M$, $\mathbf{Z}_\mathbf{X}^{(i)}$ and $\mathbf{Z}_\mathbf{Y}^{(i)}$ represent short \emph{temporally coherent} windows presumed to correspond to the \emph{same} local dynamics. To encourage local manifold consistency, we propose a unified \emph{local latent manifold alignment} (LLMA) objective that enforces both \emph{first-order} and \emph{second-order} consistency in the latent space.

\paragraph{First-order (sequence-contrastive) alignment.}
To align the latent representations of corresponding time windows, we adopt a contrastive loss \cite{oord2018representation} adapted to the temporal structure of the latent space by \emph{pulling} together time-matched local latent subsequences, $(\mathbf{Z}_\mathbf{X}^{(i)}, \mathbf{Z}_\mathbf{Y}^{(i)})$, from the two modalities and \emph{pushing} apart time-mismatched pairs from the same sequence, $(\mathbf{Z}_\mathbf{X}^{(i)}, \mathbf{Z}_\mathbf{Y}^{(j)}) \forall i\neq j$, as well as pairs
from different sequences in a batch. Formally, for $M$ windows, 
we define:
\begin{equation}
\begin{aligned}
&\mathcal{L}_{\mathrm{contrast}}=\\
    &-\frac{1}{M}
    \sum_{i=1}^{M}\log
    \frac{\exp\bigl({\mathrm{sim}(\mathbf{Z}_\mathbf{X}^{(i)}, \mathbf{Z}_\mathbf{Y}^{(i)})}/{\tau}\bigr)}{
        \sum_{j}\exp\bigl({\mathrm{sim}(\mathbf{Z}_\mathbf{X}^{(i)}, \mathbf{Z}_\mathbf{Y}^{(j)})}/{\tau}\bigr)
        + \sum_{\text{other seq}}(\cdot)
    },
    \end{aligned}
    \label{eq:l_contrast}
\end{equation}

where $\mathrm{sim}(.)$ represents a similarity function such as dot product or cosine similarity and $\tau$ is a temperature parameter. By locally aligning short-term dynamics, the model ensures that the local neighborhoods in the latent spaces derived from the two modalities reflect the same underlying state in each window.


\paragraph{Second-order (covariance) alignment.} Beyond the pairwise similarity of the local latent manifold, we also align their internal structure, using a covariance alignment term that enforces the observation streams exhibit similar second-order statistics in their latent space. 
For each time step \(l\), let \(\mathbf{Z}_\mathbf{X}^{(l)}\) and \(\mathbf{Z}_\mathbf{Y}^{(l)}\) denote the corresponding latent vectors for the two modalities. We compute the covariance matrices of these vectors (in a local neighborhood or across the entire sequence) and match them via:
\begin{equation}
    \mathcal{L}_{\mathrm{cov}}
    =
    \frac{1}{L}\,\sum_{l=1}^L 
    \mathrm{MSE}\!\Bigl(\mathrm{cov}\bigl(\mathbf{Z}_\mathbf{X}^{(l)}\bigr),\;\mathrm{cov}\bigl(\mathbf{Z}_\mathbf{Y}^{(l)}\bigr)\Bigr).
    \label{eq:l_cov}
\end{equation}

By matching the covariance matrices of $\textbf{Z}_\textbf{X}$ and $\textbf{Z}_\textbf{Y}$, we encourage both views to represent the same local manifold shape and correlation structure among latent dimensions, preserving the system’s fundamental coupling and synergy patterns. 

Finally, we form a single local latent manifold alignment (LLMA) loss by combining these two alignment components:
\begin{equation}
    \mathcal{L}_{\mathrm{LLMA}}
    =
    \mathcal{L}_{\mathrm{contrast}}
    \;+\;
    \mathcal{L}_{\mathrm{cov}}.
    \label{eq:llma}
\end{equation}

\paragraph{Overall learning objective.}
In addition to the local latent alignment, we use standard denoising objectives for each modality. Let $\mathcal{L}_{\mathrm{denoise}}^\mathbf{X}(\theta)$ and $\mathcal{L}_{\mathrm{denoise}}^\mathbf{Y}(\phi)$ be the respective mean-squared error losses for generating $\mathbf{X}_0$ and $\mathbf{Y}_0$ from their noisy versions. 
Furthermore, we apply an energy conservation term $\mathcal{L}_{\mathrm{energy}}(\theta, \phi)$ that penalizes large deviations from expected energy levels in $\mathbf{X}_0$ and/or $\mathbf{Y}_0$.


Altogether, the \emph{joint} objective combines:
\begin{equation}
\label{eq:overall_loss}
\begin{split}
\mathcal{L}(\theta, \phi)
\;=\;&
\mathcal{L}_{\mathrm{denoise}}^\mathbf{X}(\theta)
\;+\;
\mathcal{L}_{\mathrm{denoise}}^\mathbf{Y}(\phi)
\;+\; 
\alpha\,\mathcal{L}_{\mathrm{LLMA}}(\theta, \phi)
\\
&\;+\;
\gamma\,(\mathcal{L}_{\mathrm{energy}}^\mathbf{X}(\theta)
\;+\;
\mathcal{L}_{\mathrm{energy}}^\mathbf{Y}(\phi)), 
\end{split}
\end{equation}

where $\alpha$ and $\gamma$ are weighting coefficients for the local latent manifold alignment and energy constraints, respectively. We design $\alpha$ to be a learned parameter and set $\gamma$ to 1, based on experiments. By minimizing~\eqref{eq:overall_loss}, the framework \textbf{(i)} learns to denoise each modality given partial information from the other (mutual conditioning), \textbf{(ii)} aligns local embeddings to ensure consistency of short-term latent dynamics, and \textbf{(iii)} penalizes discrepancies in the energy levels between generated and real trajectories.



\subsection{Proposed algorithm}

We outline the training procedure in Algorithm~\ref{alg:training}. The algorithm integrates the conditional diffusion process with latent alignment.

\begin{algorithm}[ht!]
\caption{Training Mutually-Aligned Diffusion with Local Latent Manifold Alignment}
\label{alg:training}
\DontPrintSemicolon
\SetAlgoLined

\KwIn{Paired datasets $(\mathbf{X}, \mathbf{Y})$, noise schedule $(\alpha_t)_{t=1}^{T'}$, 
      alignment weight $\lambda$, batch size $B$, sequence length $T$}
\KwOut{Learned parameters $\theta,\phi$ for $p(\mathbf{X}|\mathbf{Y})$ and $p(\mathbf{Y}|\mathbf{X})$}

\textbf{Initialize} $\theta,\phi$ and optimizers (e.g., AdamW).\\
\For{$\text{epoch} = 1 \dots N_{\text{epochs}}$}{
  \ForEach{\text{batch} $(\mathbf{X}_0, \mathbf{Y}_0)$ \text{of size} $B$}{
    \textbf{1. Noisy Inputs:}\\
    Sample $t \sim \mathrm{Uniform}\{1,\dots,T'\}$;\quad 
    $\bm{\epsilon}_X, \bm{\epsilon}_Y \sim \mathcal{N}(\mathbf{0},\mathbf{I})$;\\
    $\mathbf{X}_t \gets \sqrt{\beta_t}\,\mathbf{X}_0 \;+\;\sqrt{1-\beta_t}\,\bm{\epsilon}_X;$\;
    $\mathbf{Y}_t \gets \sqrt{\beta_t}\,\mathbf{Y}_0 \;+\;\sqrt{1-\beta_t}\,\bm{\epsilon}_Y;$

    \textbf{2. Denoising Predictions:}\\
    $\hat{\mathbf{X}}_0 \gets p_{\theta}(\mathbf{X}_t,\mathbf{Y},t), \quad 
     \hat{\mathbf{Y}}_0 \gets p_{\phi}(\mathbf{Y}_t,\mathbf{X},t).$

    \textbf{3. Denoising Objective:}\\
    $\mathcal{L}_{\mathrm{denoise},X} = \|\mathbf{X}_0 - \hat{\mathbf{X}}_0\|^2,\quad
     \mathcal{L}_{\mathrm{denoise},Y} = \|\mathbf{Y}_0 - \hat{\mathbf{Y}}_0\|^2.$

    \textbf{4. Energy Conservation Objective:}\\
    $E(\mathbf{X}) \!=\! \tfrac{1}{2}m(\nabla_l \mathbf{X})^2 \,(m=1);$ \quad
    $\mathcal{L}_{\mathrm{energy},X} = \|E(\mathbf{X}_0) - E(\hat{\mathbf{X}}_0)\|^2,$\;
    $\mathcal{L}_{\mathrm{energy},Y} = \|E(\mathbf{Y}_0) - E(\hat{\mathbf{Y}}_0)\|^2.$

    \textbf{5. Latent Extraction \& Alignment:}\\
    $\mathbf{Z}_X \gets h_X(\mathbf{X}_t, t),\;\;\mathbf{Z}_Y \gets h_Y(\mathbf{Y}_t, t).$\\
    \emph{(Subdivide $\mathbf{Z}_X, \mathbf{Z}_Y$ into local subsequences, each of length $C$.)}\\
    $\mathcal{L}_{\mathrm{LLMA}} = \mathcal{L}_{\mathrm{contrast}}(\mathbf{Z}_X,\mathbf{Z}_Y)
      \,+\,\mathcal{L}_{\mathrm{cov}}(\mathbf{Z}_X,\mathbf{Z}_Y).$

    \textbf{6. Combine Objectives:}\\
    $\mathcal{L}_{\mathrm{total}}(\theta,\phi) = \mathcal{L}_{\mathrm{denoise},X} + \mathcal{L}_{\mathrm{denoise},Y} \;+\; 
      \alpha\,\mathcal{L}_{\mathrm{LLMA}} \;+\; 
      \gamma\,(\mathcal{L}_{\mathrm{energy},X} + \mathcal{L}_{\mathrm{energy},Y}).$

    \textbf{7. Update Parameters:}\\
    $\theta \gets \theta - \lambda_\theta\, \nabla_\theta \mathcal{L}_{\mathrm{total}},\quad
     \phi \gets \phi - \lambda_\phi\, \nabla_\phi \mathcal{L}_{\mathrm{total}}.$
  }
}
\Return $\theta, \phi$
\end{algorithm}

\section{Experiments}


\subsection{Dataset}


The evaluation is conducted on multiple biomechanical modalities, capturing complementary signals that collectively describe the locomotor process \cite{embry2018effect}. 
These datasets represent a broad range of locomotor conditions relevant to the domain, recorded from a continuum of gait tasks, comprising approximately 1,540,000 samples collected across 27 distinct locomotion profiles from ten subjects. These profiles include variations in walking speeds (0.8 to 1.2 m/s) and inclines, ranging from steep declines (-10°) to steep inclines (+10°) in increments of 2.5°. Importantly, the dataset captures not only steady-state locomotion but also transitions between different locomotor conditions, providing a comprehensive representation of human movement variability. Data collection were performed using motion capture system, which recorded multimodal information, including precise time-varying joint kinematics, joint kinetics, and ground reaction forces. This dataset offers a diverse and realistic foundation for evaluating the proposed method, ensuring its relevance and robustness within the field of biomechanics.


\subsection{Evaluation Details}

We evaluate our model on different cross‐modal biomechanical observations, namely, angles‐moments, moments‐ground reaction forces (GRFs), and angles‐GRFs, using data obtained from time‐varying joint kinematics, body kinematics, joint kinetics, and force plate recordings. We considered a temporal segment length of ${L=300}$, corresponding to two continuous gait cycles. We consider this time window such that the model can learn the transitions between different gait cycles. To assess performance, each model variant is trained and tested under multiple train-test configurations, similar to a leave-k-out cross-validation scheme that excludes different participants and task profiles at each iteration, rather than relying on one predefined test set. 
The test subsets include approx. 32,000 observations from 27 distinct task profiles, each reflecting varied speeds and conditions from a new user not seen during training. To smooth out batch-to-batch noise within an epoch during training, we take a local exponential moving average (EMA) of parameters across batches, and we finally report aggregated results across all partitions and task profiles. 

\subsection{Metrics}

To measure the generation quality, of the aligned and the unaligned models, we adopt metrics that evaluate the pointwise fidelity, distribution‐level realism, temporal structure, and representational richness. 
To evaluate the pointwise fidelity, we consider the \emph{mean-squared error (MSE)} between the generated data and the physical observations (ground truth). The ground truth signal for the reconstructed data is defined as the temporal counterpart of the conditioning data. The distribution-level realism of generated data is measured using \emph{Fréchet Inception Distance (FID)} which computes how closely the distribution of generated trajectories and the real signals match \cite{yu2021frechet, soloveitchik2021conditional} (lower the better). The \emph{predictive score} measures whether the generated data follows the same temporal dynamics as the real data by computing the error of a model trained on generated data in predicting future values of ground truth data \cite{yoon2019time}. Representational richness is measured by evaluating the performance of learned representations on a downstream classification task. 

\subsection{Model architecture}

Each diffusion model consists of four modules: 1) an input encoder,  that encodes the noise input, designed as a transformer-based encoder with four layers and a model dimension of 128, 2) a condition embedder, which encodes the guiding signal, 3) a timestep embedder, that encodes the diffusion timestep $t$, designed as a multilayer perceptron (MLP) with SiLU \cite{wang2018additive} activation, and 4) an output decoder, that generates the output at each diffusion timestep, designed as a transformer decoder with four layers. At each diffusion timestep, the noise input is linearly projected from the input space to the model space and combined with a positional and time embedding, before it passes through the encoder. At the decoder, cross-attention is computed between the condition embedding combined with positional and time embedding and the encoded noise input. The generated output is linearly projected onto the output space. 

\subsection{Effect of latent alignment on cross-modal generation}

\begin{table*}[!h]
\centering
\resizebox{0.7\textwidth}{!}{%
\begin{tabular}{@{}cc|cc|cc|cc@{}}
\toprule
\multicolumn{2}{c|}{} &
  \multicolumn{2}{c|}{} &
  \multicolumn{2}{c|}{} &
  \multicolumn{2}{c}{} \\
\multicolumn{2}{c|}{\multirow{-2}{*}{\textbf{}}} &
  \multicolumn{2}{c|}{\multirow{-2}{*}{\textbf{MSE} $\downarrow$}} &
  \multicolumn{2}{c|}{\multirow{-2}{*}{\textbf{FID} $\downarrow$}} &
  \multicolumn{2}{c}{\multirow{-2}{*}{\textbf{Pred} $\downarrow$}} \\ \midrule
 &
   &
   &
  \cellcolor[HTML]{CCFFCC} &
   &
  \cellcolor[HTML]{CCFFCC} &
   &
  \cellcolor[HTML]{CCFFCC} \\
\multirow{-2}{*}{\textbf{Modality   pair}} &
  \multirow{-2}{*}{\textbf{Direction}} &
  \multirow{-2}{*}{\textbf{w/o align}} &
  \multirow{-2}{*}{\cellcolor[HTML]{CCFFCC}\textbf{w/ align}} &
  \multirow{-2}{*}{\textbf{w/o align}} &
  \multirow{-2}{*}{\cellcolor[HTML]{CCFFCC}\textbf{w/ align}} &
  \multirow{-2}{*}{\textbf{w/o align}} &
  \multirow{-2}{*}{\cellcolor[HTML]{CCFFCC}\textbf{w/ align}} \\ \midrule
 &
   &
   &
  \cellcolor[HTML]{CCFFCC} &
   &
  \cellcolor[HTML]{CCFFCC} &
   &
  \cellcolor[HTML]{CCFFCC} \\
 &
  \multirow{-2}{*}{$X|Y$} &
  \multirow{-2}{*}{0.18±0.03} &
  \multirow{-2}{*}{\cellcolor[HTML]{CCFFCC}\textbf{0.14±0.02}} &
  \multirow{-2}{*}{37.8±9.6} &
  \multirow{-2}{*}{\cellcolor[HTML]{CCFFCC}\textbf{32.4±7.2}} &
  \multirow{-2}{*}{0.18±0.06} &
  \multirow{-2}{*}{\cellcolor[HTML]{CCFFCC}\textbf{0.16±0.03}} \\
\multirow{-4}{*}{\textbf{angles--moments}} &
  \multirow{-2}{*}{$Y|X$} &
  \multirow{-2}{*}{0.08±0.02} &
  \multirow{-2}{*}{\cellcolor[HTML]{CCFFCC}\textbf{0.07±0.01}} &
  \multirow{-2}{*}{20.4±12.1} &
  \multirow{-2}{*}{\cellcolor[HTML]{CCFFCC}\textbf{14.2±2.8}} &
  \multirow{-2}{*}{0.08±0.01} &
  \multirow{-2}{*}{\cellcolor[HTML]{CCFFCC}\textbf{0.07±0.01}} \\
 &
   &
   &
  \cellcolor[HTML]{CCFFCC} &
   &
  \cellcolor[HTML]{CCFFCC} &
   &
  \cellcolor[HTML]{CCFFCC} \\
 &
  \multirow{-2}{*}{$X|Y$} &
  \multirow{-2}{*}{0.22±0.03} &
  \multirow{-2}{*}{\cellcolor[HTML]{CCFFCC}\textbf{0.19±0.03}} &
  \multirow{-2}{*}{66.7±51.5} &
  \multirow{-2}{*}{\cellcolor[HTML]{CCFFCC}\textbf{40.4±8.3}} &
  \multirow{-2}{*}{0.30±0.23} &
  \multirow{-2}{*}{\cellcolor[HTML]{CCFFCC}\textbf{0.16±0.02}} \\
\multirow{-4}{*}{\textbf{angles--GRF}} &
  \multirow{-2}{*}{$Y|X$} &
  \multirow{-2}{*}{0.07±0.03} &
  \multirow{-2}{*}{\cellcolor[HTML]{CCFFCC}\textbf{0.06±0.03}} &
  \multirow{-2}{*}{24.8±34.2} &
  \multirow{-2}{*}{\cellcolor[HTML]{CCFFCC}\textbf{5.8±3.6}} &
  \multirow{-2}{*}{0.12±0.12} &
  \multirow{-2}{*}{\cellcolor[HTML]{CCFFCC}\textbf{0.08±0.07}} \\
 &
   &
   &
  \cellcolor[HTML]{CCFFCC} &
   &
  \cellcolor[HTML]{CCFFCC} &
   &
  \cellcolor[HTML]{CCFFCC} \\
 &
  \multirow{-2}{*}{$X|Y$} &
  \multirow{-2}{*}{0.08±0.02} &
  \multirow{-2}{*}{\cellcolor[HTML]{CCFFCC}\textbf{0.07±0.02}} &
  \multirow{-2}{*}{16.5±4.0} &
  \multirow{-2}{*}{\cellcolor[HTML]{CCFFCC}\textbf{13.7±3.2}} &
  \multirow{-2}{*}{0.08±0.01} &
  \multirow{-2}{*}{\cellcolor[HTML]{CCFFCC}\textbf{0.07±0.01}} \\
\multirow{-4}{*}{\textbf{moments--GRF}} &
  \multirow{-2}{*}{$Y|X$} &
  \multirow{-2}{*}{0.03±0.02} &
  \multirow{-2}{*}{\cellcolor[HTML]{CCFFCC}\textbf{0.03±0.02}} &
  \multirow{-2}{*}{6.6±2.5} &
  \multirow{-2}{*}{\cellcolor[HTML]{CCFFCC}\textbf{4.3±2.5}} &
  \multirow{-2}{*}{0.07±0.04} &
  \multirow{-2}{*}{\cellcolor[HTML]{CCFFCC}\textbf{0.05±0.04}} \\ \bottomrule
\end{tabular}%
}
\caption{Comparison of cross-modal generation performance of mutually aligned diffusion models for each modality pair, trained with and without latent alignment. The performance is evaluated using the discrepancy (MSE) between generated and ground truth trajectories, Fréchet Inception Distance (FID), and predictive score (predictive error), all of whose lower values indicate better performance. Training with latent alignment improves cross-modal generation quality across all modalities tested under all the different metrics evaluated here. }
\label{tab:metrics}
\end{table*}

\begin{figure}
    \centering
    \includegraphics[width=0.6\linewidth]{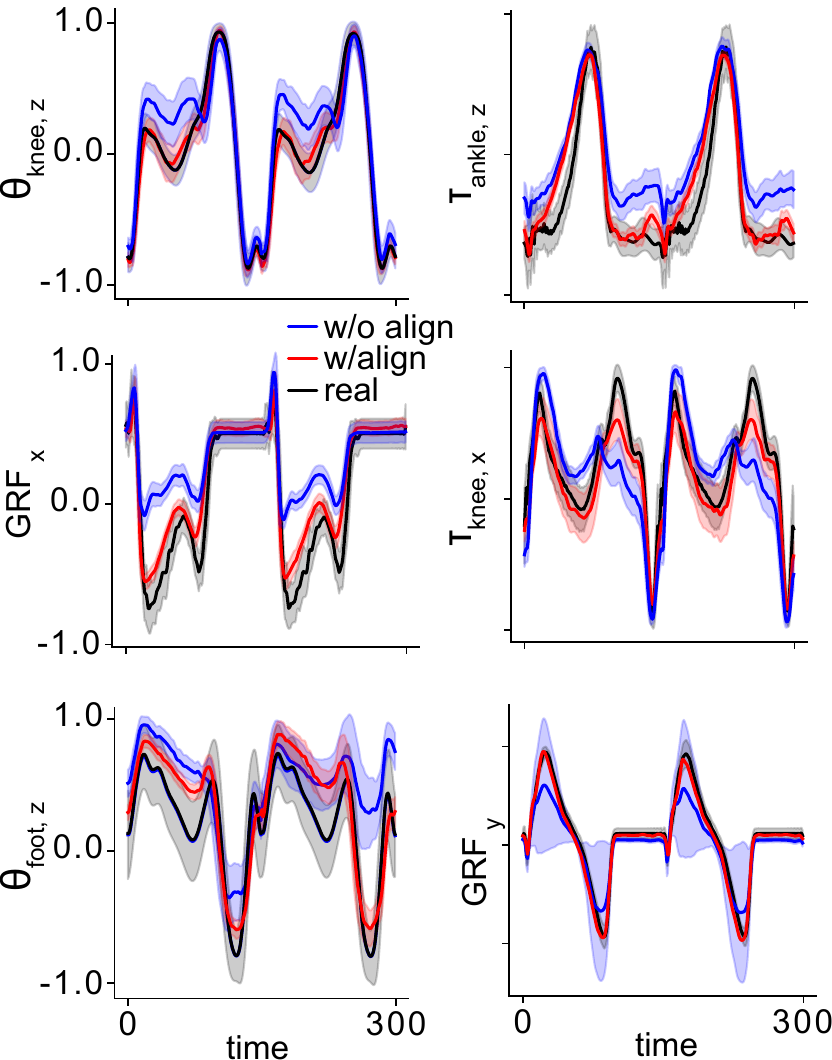} 
    \caption{Comparison of real and generated trajectories using models trained with and without latent alignment of diffusion models. Latent alignment improves the quality of generated samples. The shaded region represents the standard deviation. }
    \label{fig:reconstructions}
\end{figure}

The main proposition of our work is that by simply aligning the latent space of two independent diffusion models that generate one modality conditioned on another, we can improve cross-modal generation performance. To evaluate this proposition, we first analyzed whether the alignment of the latent embeddings from the separate models $p_\theta(\mathbf{X}|\mathbf{Y})$ and $p_\phi(\mathbf{Y}|\mathbf{X})$ that learn to generate each modality can improve the quality of their generated gait trajectories using different metrics such as MSE, FID, and predictive score. We found that latent alignment through local latent manifold alignment (LLMA) improves the cross-modal generation accuracy for all the different modalities tested (Tab. \ref{tab:metrics}). This was further illustrated by a better agreement of the generated trajectories from the aligned models of the different modalities with the ground truth trajectories (Fig. \ref{fig:reconstructions})

\subsection{Comparison with other alignment methods}

\begin{table*}[!h]
\centering
\resizebox{0.7\textwidth}{!}{%
\begin{tabular}{@{}lcclcclcc@{}}
\toprule
\multicolumn{9}{c}{\textbf{Linear probing} $\uparrow$}                                                      \\ \midrule
\multicolumn{1}{r}{Modality pair} &
  \multicolumn{2}{c}{\textbf{Angles -- Moments}} &
   &
  \multicolumn{2}{c}{\textbf{Angles -- GRF}} &
   &
  \multicolumn{2}{c}{\textbf{Moments -- GRF}} \\ \cmidrule(lr){2-3} \cmidrule(lr){5-6} \cmidrule(l){8-9} 
\multicolumn{1}{l}{Alignment} & $X\mid Y$     & $Y\mid X$     &  & $X\mid Y$     & $Y\mid X$     &  & $X\mid Y$     & $Y\mid X$     \\ \midrule \noalign{\vspace{3pt}} 
Barlow & 0.70±0.08 & 0.71±0.08 &  & 0.64±0.06 & 0.63±0.06 &  & 0.53±0.08 & 0.51±0.10 \\ \noalign{\vspace{3pt}} 
SimCLR &
  {\ul 0.82\textpm 0.04} &
  \textbf{0.79±0.06} &
   &
  {\ul 0.68±0.06} &
  \textbf{0.78±0.04} &
   &
  0.78±0.04 &
  {0.80±0.08} \\ \noalign{\vspace{3pt}} 
MSE &
  0.72±0.04 &
  0.72±0.06 &
   &
  0.62±0.10 &
  {0.74±0.03} &
   &
  \textbf{0.82±0.04} &
  \textbf{0.83±0.05} \\ \noalign{\vspace{3pt}} 
VICReg & 0.65±0.09 & 0.64±0.06 &  & 0.66±0.07 & 0.62±0.07 &  & 0.54±0.07 & 0.59±0.09 \\ \noalign{\vspace{3pt}} 
\rowcolor[HTML]{CCFFCC} 
LLMA &
  \textbf{0.86±0.05} &
  {\ul 0.78±0.05} &
   &
  \textbf{0.80±0.06} &
  {\ul 0.75±0.04} &
   &
  \textbf{0.82±0.04} &
  \textbf{0.83±0.07} \\ \midrule
\multicolumn{9}{c}{\textbf{Nonlinear probing} $\uparrow$}                                                   \\ \midrule \noalign{\vspace{3pt}} 
Barlow & 0.72±0.06 & 0.73±0.07 &  & 0.66±0.08 & 0.68±0.05 &  & 0.63±0.07 & 0.57±0.10 \\ \noalign{\vspace{3pt}} 
SimCLR &
  {\ul 0.83±0.05} &
  \textbf{0.80±0.06} &
  &
  {\ul 0.74±0.07} &
  \textbf{0.81±0.05} &
   &
  0.64±0.07 &
  0.68±0.09 \\ \noalign{\vspace{3pt}} 
MSE & 0.74±0.05 & 0.75±0.06 &  & 0.65±0.10 & 0.76±0.05 &  & \textbf{0.85±0.05} & \textbf{0.85±0.05} \\ \noalign{\vspace{3pt}} 
VICReg & 0.64±0.09 & 0.64±0.07 &  & 0.72±0.07 & 0.66±0.06 &  & 0.64±0.07 & 0.68±0.09 \\ \noalign{\vspace{3pt}} 
\rowcolor[HTML]{CCFFCC} 
LLMA &
  \textbf{0.86±0.05} &
  \textbf{0.80±0.05} &
  \textbf{} &
  \textbf{0.83±0.06} &
  {\ul 0.78±0.05} &
   &
  \textbf{0.85±0.05} &
  {\ul 0.84±0.05} \\ \bottomrule
\end{tabular}%
}
\caption{Quality of learned representations of different latent alignment methods quantified as the performance on locomotion profile classification using linear and nonlinear probes (mean and standard deviation across test sets, bold indicates best performing and underline indicates second best performing). Local latent manifold alignment (LLMA) outperforms state-of-the-art self-supervised methods across four out of six modalities tested, and performed second best in the remaining two modalities. 
} 
\label{tab:probing}
\end{table*}

For aligning the latent embeddings, we propose LLMA (local latent manifold alignment), which combines 1) a first-order alignment objective, $\mathcal{L}_{\mathrm{contrast}}$, represented as a modified version contrastive objective for time series data, 
and 2) a second-order alignment objective, $\mathcal{L}_{\mathrm{cov}}$, represented as the deviation between the covariance of the latent dimensions at each time step. We evaluated the quality of the latent representations learned by LLMA by comparing it with state-of-the-art self-supervised learning methods, such as SIMCLR \cite{chen2020simple}, Barlow twins \cite{zbontar2021barlow}, and VICREG \cite{bardes2021vicreg}, that work based on latent space alignment. Additionally, a simple baseline with mean-squared error between the latents of the two models was also evaluated. The comparison was done by evaluating the performance on a downstream task. We used the classification of the locomotion task label as the downstream task. Each input sample from either modality $X$ or $Y$ belongs to one of the 27 locomotion tasks defined by the distinct speed of the subject and incline of the walking ground. A linear or nonlinear classification model was trained to predict the locomotion task label from the encoder outputs at the final diffusion step. A higher linear/nonlinear probing score would indicate better discrimination of the underlying locomotion tasks in the latent space, which suggests a better quality of the learned representations. We found that LLMA outperformed the state-of-the-art alignment methods in the downstream tasks for four out of six models tested and performed second best for the remaining two models (Tab. \ref{tab:probing}). 

\subsection{Effect of alignment on the learned representations}

\begin{figure}[!thbp]
    \centering
    \includegraphics[width=0.6\linewidth]{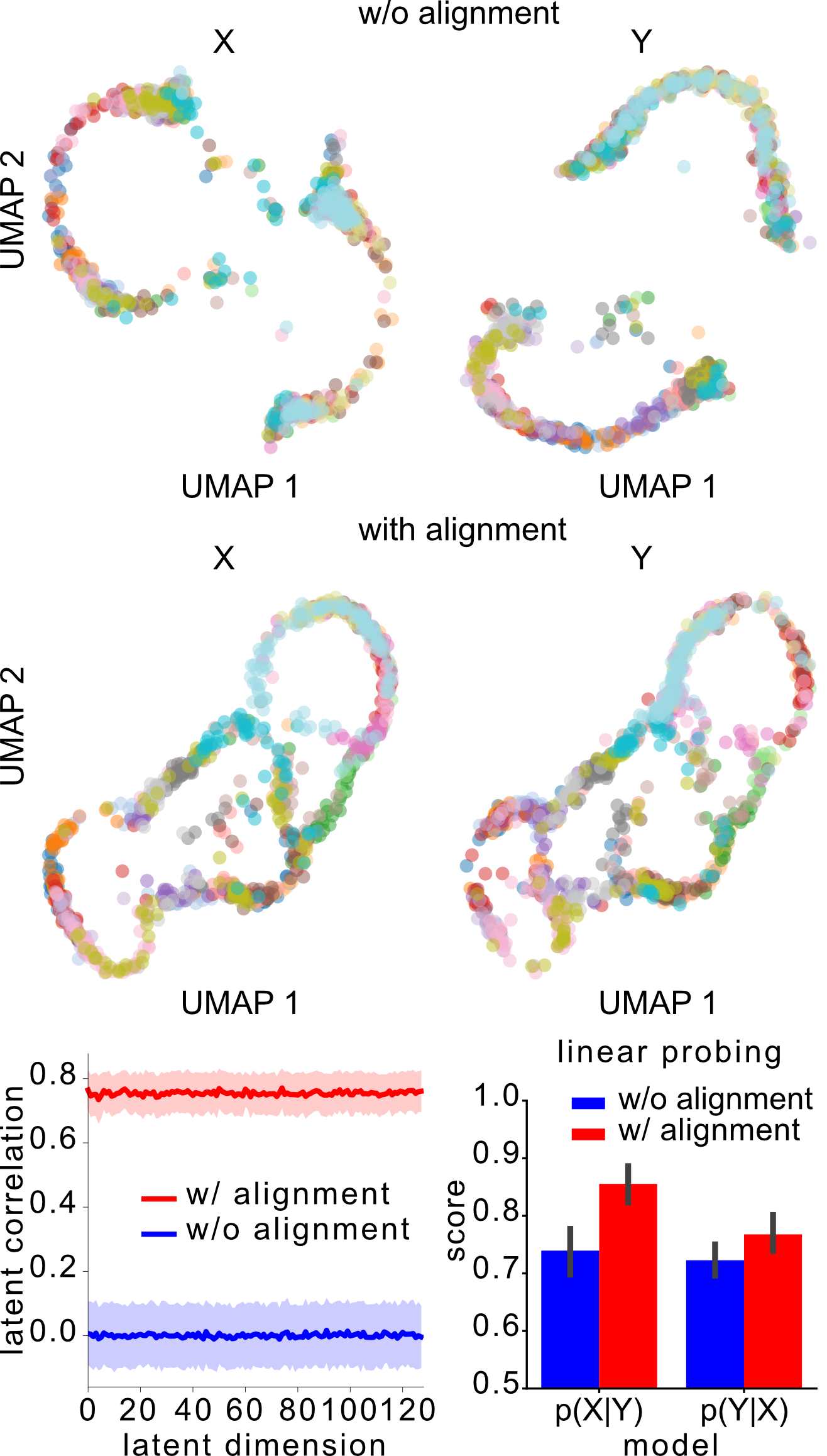}
    \caption{(Top) Visualization of latent embeddings of the models $p(X|Y)$ and $p(Y|X)$ on a held-out subject data trained without and with latent alignment. The samples are color-coded by locomotion task label. The latent representations of the models trained without alignment show high modality-specific separation in latent space. On the other hand, latent representations of the models trained with alignment show a merge of the latent spaces for both modalities with samples belonging to the same task occupying overlapping subspaces. (Bottom left) The correlation between latent representations of the two modality-specific models on held-out test data. Models trained with alignment show a high correlation between the latent spaces of the two modalities. (Bottom right)  Performance of a linear classifier in discriminating the locomotion tasks from the latent representation of modality-specific models. Latent representations from models trained with alignment give better accuracy emphasizing a clearer separation of tasks in their latent spaces. 
    }    
    \label{fig:latent_visualization}
\end{figure}

Our idea of latent alignment is motivated by the assumption that the different modalities arise from the same underlying system. Thus, aligning the two views can encourage the latent representations of $X$ and $Y$ to capture shared or complementary information, thus learning better representations than each modality can learn individually. 
To test this hypothesis, we evaluated how the latent alignment influences the learned representations of the model. We first visualized the latent space of the two models learned without and with alignment using UMAP \cite{mcinnes2018umap}. The latent space of models trained with alignment showed a high correlation to each other, with samples belonging to the same task occupying overlapping subspaces (Fig. \ref{fig:latent_visualization}). This was further illustrated by the superior performance of the aligned models on the downstream linear classification of the locomotion task profiles. Thus, we found that, by simply aligning the latent spaces of two modalities, it is possible to enhance the representational quality of individual modalities.  This is not an effect of mutual conditioning at the decoder stage or due to the energy conservation objective since non-aligned models were also trained by including these factors.

\begin{table}[!h]
\centering
\resizebox{0.5\textwidth}{!}{%
\begin{tabular}{@{}l|ccc|cc@{}}
\toprule
\textbf{} & $\mathcal{L}_{\mathrm{energy}}$ & $\mathcal{L}_{\mathrm{contrast}}$ & $\mathcal{L}_{\mathrm{cov}}$ & $X|Y$ & $Y|X$ \\ \midrule
LLMA w/o $\mathcal{L}_{\mathrm{contrast}}$ & \checkmark &            & \checkmark & 0.18±0.03          & 0.08±0.03          \\
LLMA w/o $\mathcal{L}_{\mathrm{cov}}$      & \checkmark & \checkmark &            & 0.17±0.03          & 0.07±0.02          \\
LLMA w/o $\mathcal{L}_{\mathrm{energy}}$   &            & \checkmark & \checkmark & 0.17±0.02          & 0.07±0.02          \\
\rowcolor[HTML]{CCFFCC} 
\textbf{LLMA}                              & \checkmark & \checkmark & \checkmark & \textbf{0.14±0.02} & \textbf{0.07±0.01} \\ \bottomrule
\end{tabular}%
}
\caption{Effect of ablation of individual components of the objective on the model performance measured using MSE (Mean and standard deviation across test sets; lower the better). Removing each component worsens the model's cross-modal generation capability, whereas all the components together are required to achieve the best performance. } 
\label{tab:ablations}
\end{table}

\subsection{Ablations}



Finally, we conducted an ablation study to assess the contribution of each component in our overall loss term (Eq. \ref{eq:overall_loss}). Specifically, we removed the energy conservation objectives ($\mathcal{L}_{energy, \mathrm{X}}$ and $\mathcal{L}_{energy, \mathrm{Y}}$), the covariance alignment objective ($\mathcal{L}_{\mathrm{cov}}$), and the contrastive alignment objective ($\mathcal{L}_{\mathrm{constrast}}$), individually, and compared these variants against the full, non-ablated objective. Our results show that each component is necessary for achieving the best performance from our method (Tab. \ref{tab:ablations}).




\section{Conclusion}


We presented a novel mutually-aligned diffusion framework for cross-modal biomechanical time-series generation, grounded in a dynamical systems perspective. By applying a local latent manifold alignment, comprising, first-order (sequence-contrastive) and second-order (covariance) alignment at each diffusion time step, our approach synthesizes realistic, cross-modal outputs, preserving biomechanically consistent relationships across modalities. Experiments show that this simple alignment strategy produces more accurate signal generation compared to baselines, and also enhances performance in downstream tasks, demonstrating its utility in both generative and discriminative contexts. Future work is required to validate the model performance when the assumption of modalities originating from a shared underlying process is violated. 



\section*{Broader Impact}

The proposed framework for mutually aligned cross-modal diffusion opens a wide range of possibilities in scenarios where one or more data streams are missing, noisy, or difficult to measure directly. In wearable assistive devices and robotics, it can infer absent or corrupted sensor inputs, such as force or torque data from more accessible modalities, thereby enhancing real-time control and reliability despite equipment constraints or sensor failure. Within the biomechanical domain, the ability to simulate perturbations in one modality and observe their repercussions in another offers powerful insights into how different aspects of locomotion co-evolve, informing the design of targeted rehabilitation protocols and sophisticated training regimens. By enabling more efficient sensor setups, the framework supports clinicians and researchers in long-term monitoring without requiring extensive instrumentation, broadening the potential for in-home rehabilitation and remote athlete performance tracking. Beyond biomechanics, the fundamental principles behind our cross-modal diffusion paradigm can be extended to other domains where interacting data streams arise from a shared dynamical process. For instance, in climate modeling, it could align or impute different types of geospatial and atmospheric measurements to refine weather or environmental forecasts. Even financial modeling could benefit from aligning time-series of economic indicators or market signals to better predict systemic interactions. 

\paragraph{Ethical Considerations. }
The ability to reconstruct missing data from alternative sources raises important questions about privacy, consent, and fairness, particularly when dealing with sensitive physiological information. These concerns underscore the need for robust regulatory frameworks and ethical practices to ensure responsible research and real-world implementations.

\bibliography{CDiff_Dyn}
\bibliographystyle{icml2025}

\newpage
\appendix
\onecolumn
\section{Appendix}



\subsection{Dataset and data modalities}
We used open-source biomechanical motion datasets \cite{embry2018effect} consisting of locomotion data collected as multiple subjects walked on an instrumented treadmill at varying speeds (0.8\,m/s, 1.0\,m/s, and 1.2\,m/s)  and inclines (-10\,\textdegree to 10\,\textdegree at 2.5\,\textdegree increments). The locomotion data was recorded using a 10-camera Vicon motion capture system, while the force plates in the treadmill recorded ground reaction forces (GRF). The processed data consists of three modalities 1) Kinematics that consists of 3D joint angles of hip, knee, and ankle, and 3D pelvis and foot angles, 2) joint kinetics that consists of 3D moments of hip, knee, and ankle, and 3) 3D ground reaction forces. The feature sets are represented as $(\theta_{\mathrm{hip, x}}, \theta_{\mathrm{hip, y}}, \theta_{\mathrm{hip, z}}, \theta_{\mathrm{knee, x}}, \theta_{\mathrm{knee, y}}, \theta_{\mathrm{knee, z}}, \theta_{\mathrm{ankle, x}}, \theta_{\mathrm{ankle, y}}, \theta_{\mathrm{ankle, z}}, \theta_{\mathrm{foot, x}}, \theta_{\mathrm{foot, x}}, \theta_{\mathrm{foot, x}}, \theta_{\mathrm{pelvis, x}}, \theta_{\mathrm{pelvis, y}}, \theta_{\mathrm{pelvis, z}})$, $(\tau_{\mathrm{hip, x}}, \tau_{\mathrm{hip, y}}, \tau_{\mathrm{hip, z}}, \tau_{\mathrm{knee, x}}, \mathrm{knee, y}, \tau_{\mathrm{knee, z}}, \tau_{\mathrm{ankle, x}}, \tau_{\mathrm{ankle, x}}, \tau_{\mathrm{ankle, x}})$, $(\mathrm{GRF_x}, \mathrm{GRF_y}, \mathrm{GRF_z})$. The features were normalized using min-max scaling prior to model training.

\subsection{Implementation Details}

We trained parallel diffusion models (DDPM), $p_\theta(\mathbf{X}|\mathbf{Y})$ and $p_\phi(\mathbf{Y}|\mathbf{X})$ for generating the two modalities $\mathbf{X}$ and $\mathbf{Y}$ conditioned on the other with latent alignment. Each model has the same architecture and consists of a transformer encoder and decoder, each with four layers. Inputs to both encoder and decoder were combined with sinusoidal position encoding and time embedding. The decoder additionally takes a conditional embedding derived from the other modality through a linear layer. Each model has \~\,25M tunable parameters. In contrast to having a single test set, we performed a K-fold cross-validation of the models by creating multiple versions of train and test datasets. The model training was done for \~\,50 epochs on an RTX4090 GPU computer which consumed \~\,10GB of GPU space and \~\,10\,hours for 10 cross-validation iterations. 

\subsection{Metrics} 

\paragraph{Mean-Squared Error (MSE)} Each diffusion model, 
$p(\mathbf{X}|\mathbf{Y})$ or $p(\mathbf{Y}|\mathbf{X})$, generates data for one modality, $\hat{\mathbf{X}}$ or $\hat{\mathbf{Y}}$, conditioned on the other. Since the modalities are time-series data that correspond to each other, this conditioned generation can be viewed as a cross-modal reconstruction task. The ground truth signal for the reconstructed data is defined as the temporal counterpart of the conditioning data. We then calculate the mean squared error (MSE) between the generated data, $\hat{\mathbf{X}}$ or $\hat{\mathbf{Y}}$ and the ground truth data for the respective modality. 

\paragraph{Fréchet Inception Distance (FID)} evaluates the quality of generated data by measuring the Fréchet distance (Wasserstein-2 distance) between the distributions of real and generated features \cite{yu2021frechet}. Originally designed for images, we adapt this metric for generated time series data by computing the distance in the temporal space. Given two Gaussian distributions, $\mathcal{N}(\mu, \Sigma)$ and $\mathcal{N}(\mu', \Sigma')$, respectively fitted to the real and generated feature representations, the FID is computed as: 

\begin{equation}
\label{eq:fid}
    \mathrm{FID} = \norm{\mu - \mu'}^2_2 + \mathrm{tr}(\Sigma + \Sigma' - 2(\Sigma\Sigma')^{\frac{1}{2}})
\end{equation}

\paragraph{Predictive score} This metric evaluates generation quality by assessing how well a model trained on generated data predicts future values in real data \cite{yoon2019time}. A sequence-to-sequence model (e.g., LSTM) is trained to predict the latter part of a time series from its initial part, and its performance on real data reflects the quality of the generated data, with lower errors indicating higher quality.

\subsection{Representation alignment methods}

\subsubsection{SimCLR}

SimCLR \cite{chen2020simple} is a contrastive learning approach that learns representations by bringing similar samples (positive pairs) closer in the latent space while pushing dissimilar ones (negative pairs) apart. It relies on a contrastive loss function, the Normalized Temperature-scaled Cross-Entropy Loss (NT-Xent loss), which is defined as:

\begin{equation}
\ell_{i,j} = -\log \frac{\exp\left({\text{sim}(\mathbf{z}_i, \mathbf{z}_j)}/{\tau}\right)}{\sum_{k=1}^{2N} {1}_{[k \neq i]} \exp\left({\text{sim}(\mathbf{z}_i, \mathbf{z}_k)}/{\tau}\right)}
\end{equation}

where $\mathbf{z}_i, \mathbf{z}_j$ are the embeddings of two samples, $\text{sim}(\mathbf{z}_i, \mathbf{z}_j) = \frac{\mathbf{z}_i \cdot \mathbf{z}_j}{\|\mathbf{z}_i\| \|\mathbf{z}_j\|}$ is the cosine similarity measure, $\tau$ is the temperature scaling parameter, and $N$ is the batch size. The total loss across a batch of size $N$ is computed as:

\begin{equation}
\mathcal{L}_{\text{SimCLR}} = \frac{1}{2N} \sum_{i=1}^{N} \left( \ell_{2i-1, 2i} + \ell_{2i, 2i-1} \right).
\end{equation}

We consider the latent embeddings of the corresponding samples of both modalities in a batch as positive pairs, and non-corresponding samples as negative pairs. 


\subsubsection{Barlow Twins}

Barlow Twins \cite{zbontar2021barlow} addresses the limitations of contrastive methods by eliminating the need for negative samples. It introduces a loss function that aligns the cross-correlation matrix of embeddings from two identical networks processing different augmentations of the same image (in our case two modalities). The objective is twofold: (1) to make the diagonal elements of this matrix approach one, ensuring invariance, and (2) to drive the off-diagonal elements towards zero, promoting redundancy reduction. This strategy effectively prevents collapse by decorrelating different dimensions of the representation space.

Given two embeddings $\mathbf{z}^A$ and $\mathbf{z}^B$ (where $A$ and $B$ are two modalities), it computes the cross-correlation matrix:

\begin{equation}
C_{ij} = \frac{1}{B} \sum_{n=1}^{N} z^A_n(i) z^B_n(j)
\end{equation}

where $N$ is the batch size and $z^{(\cdot)}(i)$ represents the $i$-th feature dimension. The Barlow Twins loss consists of two key terms:

\begin{itemize}
    \item \textbf{Invariance term}: Ensures that representations of the same input under different augmentations are similar: \(\sum_{i} (1 - C_{ii})^2.\)
    \item \textbf{Redundancy reduction term}: Enforces decorrelation across different dimensions to prevent representational collapse: \(\sum_{i \neq j} C_{ij}^2.\)
\end{itemize}

The final loss function is formulated as:

\begin{equation}
\mathcal{L}_{\text{Barlow}} = \sum_{i} (1 - C_{ii})^2 + \lambda \sum_{i \neq j} C_{ij}^2,
\end{equation}

where $\lambda$ is a balancing hyperparameter.


\subsubsection{VICReg}

VICReg (Variance-Invariance-Covariance Regularization) \cite{bardes2021vicreg} extends Barlow Twins by adding an explicit variance regularization term, preventing representational collapse through three objectives:

\begin{itemize}
    \item \textbf{Invariance}: Ensures consistency between augmented views, similar to SimCLR and Barlow Twins:
    \(
    \mathcal{L}_{\text{inv}} = \sum_{i=1}^{d} \| \mathbf{z}^A(i) - \mathbf{z}^B(i) \|^2
    \)
    where $A$ and $B$ are two modalities, and $z^{(\cdot)}(i)$ represents the $i$-th feature dimension. 
    
    \item \textbf{Variance regularization}: Ensures that the standard deviation of each embedding dimension $i$ remains above a threshold $\gamma$, preventing collapse to trivial solutions:
    \(
    \mathcal{L}_{\text{var}} = \sum_{i=1}^{d} \max(0, \gamma - \sigma(\mathbf{z}(i)))^2.
    \)
    
    \item \textbf{Covariance regularization}: Reduces redundancy between different dimensions by minimizing off-diagonal terms of the covariance matrix:
    \(
    \mathcal{L}_{\text{cov}} = \sum_{i \neq j} C_{ij}^2, \quad C = \frac{Z^\top Z}{N}, 
    \)
    where $N$ is the batch size. 
\end{itemize}

The total VICReg loss function is:

\begin{equation}
\mathcal{L}_{\text{VICReg}} = \lambda_{\text{inv}} \mathcal{L}_{\text{inv}} + \lambda_{\text{var}} \mathcal{L}_{\text{var}} + \lambda_{\text{cov}} \mathcal{L}_{\text{cov}}.
\end{equation}

This approach provides a balance between alignment and diversity constraints, ensuring that representations are meaningful, discriminative, and well-distributed.

\subsection{Further visualizations}

\begin{figure}[!h]
    \centering
    \includegraphics[width=\linewidth]{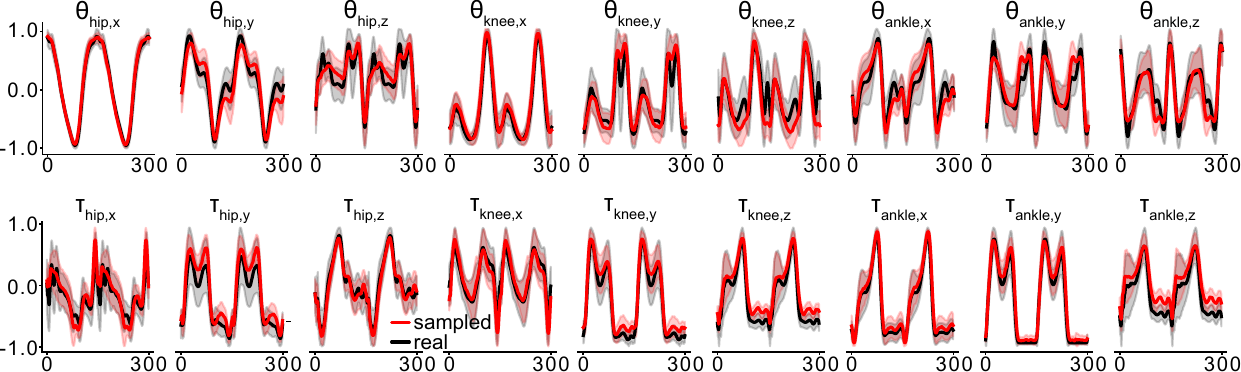}
    \caption{Real (black) and sampled (red) trajectories of joint angles (top) and joint moments (bottom) generated by latent aligned cross-modal diffusion models. All the generated trajectories follow the ground truth trajectories closely. Shaded region represents standard deviation. }
    \label{fig:reconstructions_appendix}
\end{figure}

\begin{figure}[!h]
    \centering
    \includegraphics[width=0.35\linewidth]{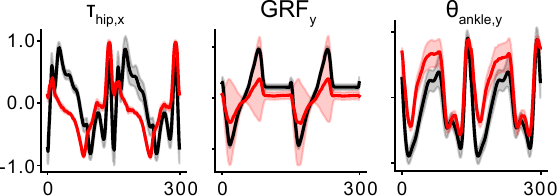}
    \caption{Example failure cases of the model for the prediction of the three modalities. Failure cases mostly occur when the underlying true signal shows high variability, or due to sign changes in the sampled signals. }
    \label{fig:failure_cases_appendix}
\end{figure}

\begin{figure}[!h]
    \centering
    \includegraphics[width=\linewidth]{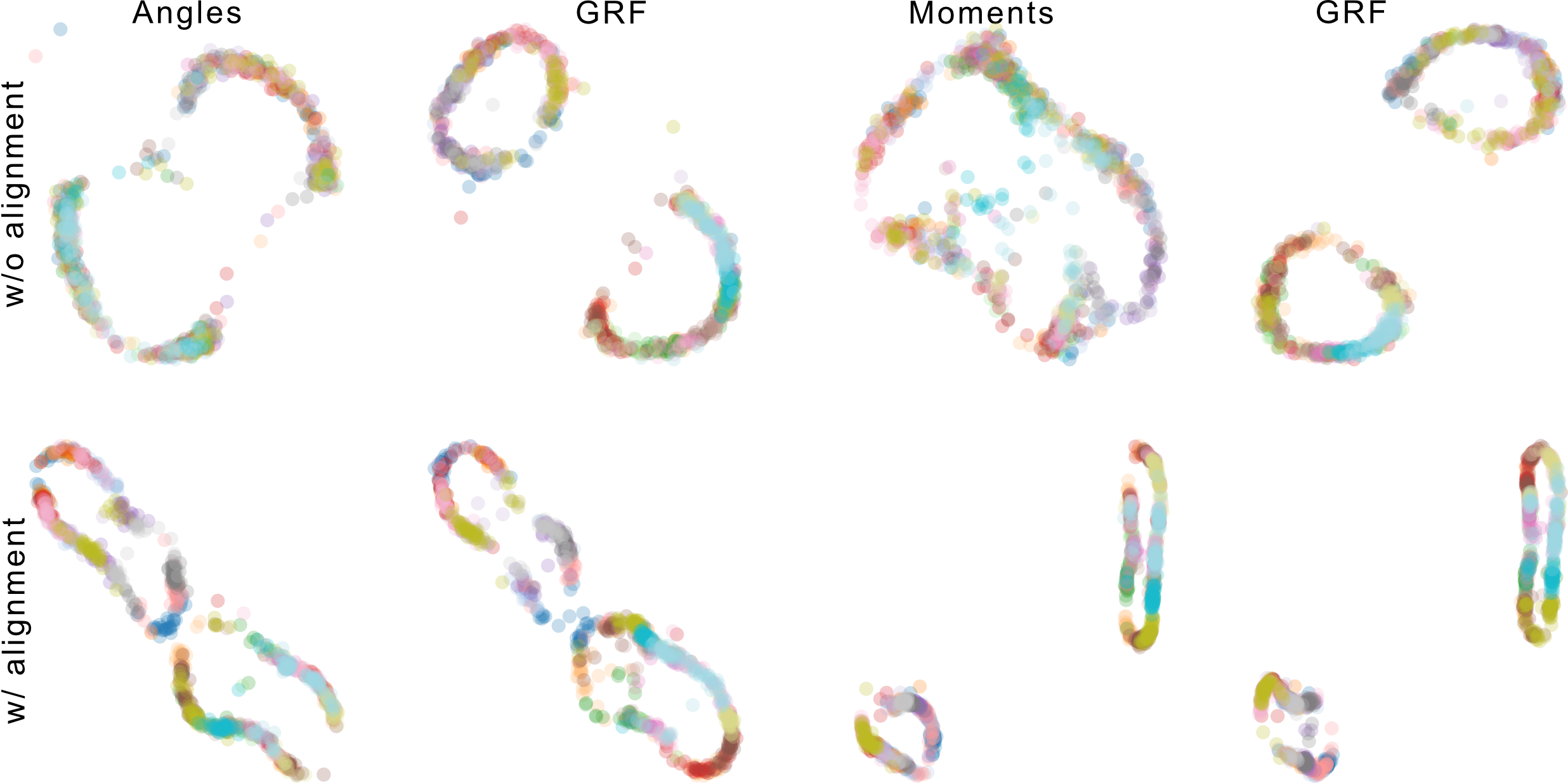}
    \caption{UMAP visualizations of latent spaces of cross-modal diffusion models for joint angles and GRF (left) and joint moments and GRF (right) for diffusion models trained independently (w/o alignment) and with latent alignment. The latent space of the latent aligned models shows a correlation in the structure and arrangement of locomotion tasks (color codes), whereas the latent space of the independently trained models shows a modality-specific structure without observable correlations. }
    \label{fig:umap_visualization_appendix}
\end{figure}

\end{document}